\pdfoutput=1
\PassOptionsToPackage{nocompress}{cite}
\RequirePackage[T1]{fontenc}
\documentclass[letterpaper,10pt,conference]{ieeeconf}
\IEEEoverridecommandlockouts
\usepackage{amsmath,amssymb,graphicx,booktabs,multirow,cite,xcolor}
\usepackage{threeparttable}
\newcommand{\metricup}[1]{#1$\uparrow$}

\usepackage{tikz}
\usetikzlibrary{arrows.meta,calc,positioning,shapes.geometric}
\usepackage{url}
\usepackage{placeins}
\usepackage{hyperref}
\usepackage{xcolor}
\makeatletter

\newcommand{\teaserfigurecaption}[1]{\begingroup\def\@captype{figure}\caption{#1}\label{fig:teaser}\endgroup}
\makeatother
\newcommand{\OA}{\mathrm{OA}}
\newcommand{\OM}{\mathrm{OM}}
\newcommand{\hist}{\mathbf{H}}
\newcommand{\ctx}{\mathbf{C}}
\newcommand{\pio}{\pi_{0.5}}
\definecolor{oiBlue}{HTML}{0072B2}
\definecolor{oiSky}{HTML}{56B4E9}
\definecolor{oiGreen}{HTML}{009E73}
\definecolor{oiOrange}{HTML}{E69F00}
\definecolor{oiVermillion}{HTML}{D55E00}
\definecolor{ink}{HTML}{1F2937}
\definecolor{muted}{HTML}{5F6B78}
\definecolor{line}{HTML}{CBD5E1}
\definecolor{softBlue}{HTML}{EAF5FB}
\definecolor{softGreen}{HTML}{EAF7F2}
\definecolor{softOrange}{HTML}{FFF6DF}
\definecolor{softGray}{HTML}{F4F6F8}
\title{\LARGE \bf I See What You Want: Decoding Hierarchical Operator Intent from Visual Teleoperation History}
\author{%
Xiangyi~Jia$^{1}$\quad Kaixian~Qu$^{1}$\quad Shengzhi~Wang$^{1,2}$\quad
Chenyu~Yang$^{1}$\quad Robert~Katzschmann$^{1}$\quad Marco~Hutter$^{1}$\\
{\normalsize $^{1}$ETH Zurich\qquad $^{2}$Chinese University of Hong Kong}%
}
\IEEEaftertitletext{%
  \begin{minipage}{\textwidth}
    \centering
    \includegraphics[width=0.82\textwidth]{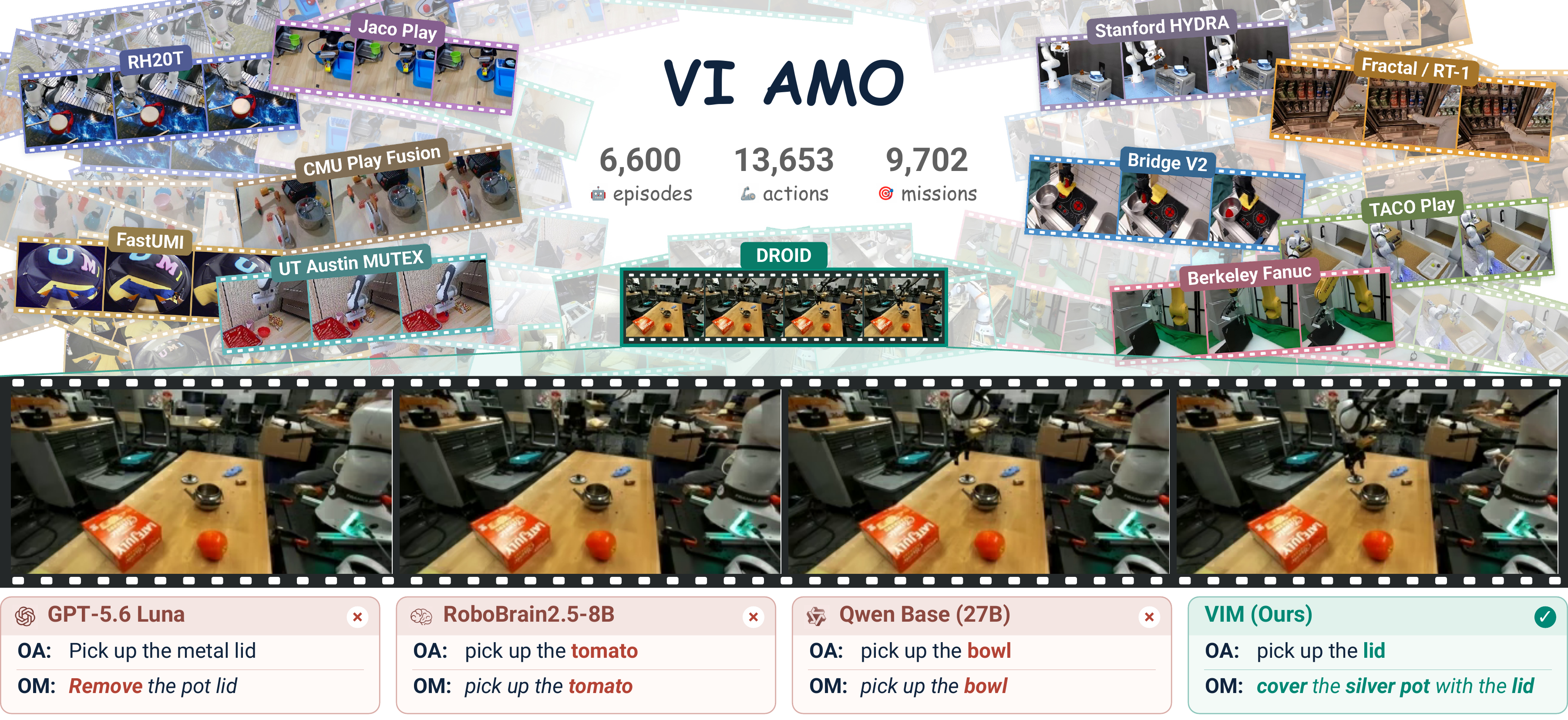}
    \teaserfigurecaption{\textbf{VI AMO and hierarchical intent prediction from VIM.} Fine-tuned on VI AMO, our proposed dataset for hierarchical intent prediction, VIM combines temporally ordered observations to jointly generate the ongoing action (OA) and the mission (OM) it serves in a single autoregressive pass. Each box shows the predicted OA on top and OM below in italics. In this example, VIM correctly predicts both intents, while the state-of-the-art frontier models fail to do so.}
  \end{minipage}%
  \vspace{2pt}%
}
\begin{document}
\bstctlcite{VIA:referenceStyle}
\maketitle
\thispagestyle{empty}
\pagestyle{empty}
\raggedbottom
\begin{abstract}

Vision-language models (VLMs) have become a powerful foundation for robotic scene understanding, task planning, and language-conditioned control. A natural question follows: can these models also decode what a human operator intends while teleoperating a robot, without any explicit instruction? We formulate this as hierarchical intent prediction, jointly inferring the ongoing atomic action and the broader mission it serves from a short visual teleoperation history. To study this, we introduce VI AMO (Vision Intention Action and Mission Ongoing), a dataset of 6,600 episodes from eleven robot sources with 13,653 temporally aligned action annotations and 9,702 mission annotations. Large proprietary models struggle with this problem: on held-out episodes, GPT-4o reaches Sentence-BERT similarities of only 0.563 and 0.470 for action and mission, and GPT-5.6 Luna achieves only 0.581 and 0.554. Fine-tuning a VLM on VI AMO yields our Vision-Intention Model (VIM), which raises these scores to 0.825 and 0.821 and still reaches 0.726 and 0.717 on the 35.4\% of test actions with both labels absent from training. Finally, as a downstream application, VIM's predicted mission enables a vision-language-action (VLA) policy to take over from the operator mid-task in a real-world environment unseen during VIM fine-tuning, reducing human teleoperation time by 34.6\% relative to GPT-4o and 39.8\% relative to GPT-5.6 Luna, while cutting intent-prediction latency by roughly 70\% on average.

\end{abstract}

\section{INTRODUCTION}
\label{sec:intro}

Vision-language models (VLMs) have emerged as a key backbone for robotic perception and reasoning. SayCan~\cite{ahn2022saycan} grounds language plans in learned
affordances, Code as Policies~\cite{liang2023code} generates executable robot
programs from language, and Inner Monologue~\cite{huang2022inner} closes the
loop with textual feedback. PragmaBot~\cite{qu2026pragmatist} uses visual
feedback and lessons from real-world experience to refine robot task plans.
On the execution side, vision-language-action
(VLA) policies map a language instruction and visual observations directly to
motor
commands~\cite{zitkovich2023rt2,kim2024openvlaopensourcevisionlanguageactionmodel,black2026pi0visionlanguageactionflowmodel,intelligence2025pi05visionlanguageactionmodelopenworld}. All of these translate human language into robot behavior. We ask the inverse
question: given only the partial visual history of a teleoperated robot, can a VLM
decode what the human operator intends, without an instruction ever being
provided?

Recent work has begun to explore this direction.
VOSA~\cite{belsare2025zeroshotuserintentrecognition} identifies candidate
target objects through detection and segmentation, and
CASPER~\cite{liu2025casper} uses VLM reasoning over teleoperation history to
select a skill--object pair from a predefined library and then executes that
skill. Both represent intent at the level of the current step, as a target object or
position, or as a skill drawn from a predefined library. However, these methods do not explicitly infer the broader mission behind the current action, yet it is what a robot needs in order to
continue: grasping an apple does not by itself reveal whether the apple should
go on a plate, to a person, or into a basket. Progress is further limited by data: existing
demonstration
corpora~\cite{walke2023bridgedata,khazatsky2024droid,openx2023}
mainly provide episode-level instructions, while existing intermediate supervision is either synthetically generated from task instructions~\cite{zawalski2024ecot}, or annotated at the primitive-skill level~\cite{chen2024rh20tp}, rather than providing temporally aligned action--mission supervision for instruction-free intent inference. 

We therefore model operator intent as a hierarchy. The \textit{ongoing
action}~(OA) is the atomic manipulation step currently being performed, such
as ``pick up the apple,'' and the \textit{ongoing mission}~(OM) is the
immediate goal it serves, such as ``put the apple on the plate.'' Both are expressed as open-vocabulary natural language, so the predicted
mission can serve as the instruction of a language-conditioned policy without
a fixed skill library. The difficulty is that a useful prediction must
arrive early: at handoff time only part of the action has been observed, so
the model must commit from a few frames and a short record of what has
already been completed.

To study this problem, we introduce VI~AMO (Vision Intention Action and
Mission Ongoing), a dataset of 6,600 episodes from eleven robot source
families with 13,653 temporally aligned action annotations and 9,702 mission
annotations. Without task-specific training, even
strong proprietary models fall short in predicting the ongoing mission: GPT-4o and GPT-5.6 Luna reach mission Sentence-BERT (SBERT) similarities of only 0.470 and 0.554. Fine-tuning an open-source VLM on VI~AMO yields the
Vision-Intention Model~(VIM), which outperforms the two proprietary VLMs, two robotics-specialized foundation models, and its untuned backbone across different evaluation metrics, achieving a mission SBERT similarity of 0.821 compared with at most 0.564 for the baselines.
On the 35.4\% of test actions whose action and mission labels are both unseen during training, VIM achieves 27.2\% higher action SBERT and 27.1\% higher mission SBERT than the strongest baseline for both action and mission prediction.

Finally, we test the utility of the inferred mission in shared autonomy, where
a robot continues a task that the operator
began~\cite{9501975,losey2022learning}. VIM's predicted mission conditions a VLA policy that takes over mid-task and completes the remaining manipulation in real-world scenarios unseen during VIM fine-tuning. With five trials per task for each
VLM condition, VIM reduces human teleoperation time by 34.6\%
relative to GPT-4o and 39.8\% relative to GPT-5.6 Luna. Each intent prediction takes approximately 0.7 s on a consumer GPU, compared with 1.9--3.5 s for the API-based baselines.

Our main contributions are as follows:
\begin{itemize}
  \item We formulate hierarchical intent prediction from visual
    teleoperation history, jointly predicting the ongoing action and the
    mission it serves as open-vocabulary language, and introduce VI~AMO, a
    dataset of 6,600 episodes with aligned action--mission annotations to
    train and evaluate it.
  \item We present VIM, a VLM fine-tuned on VI~AMO that outperforms GPT-4o,
    GPT-5.6 Luna, two embodied foundation models, and its untuned backbone on
    different metrics, reaching 0.825 and 0.821 Sentence-BERT similarity for action and
    mission on the VI~AMO test set versus at most 0.581 and 0.564 for the best
    baseline.
  \item We evaluate VIM as the intent module of a shared-autonomy system
across five real-world tasks in scenarios unseen during VIM fine-tuning.
Its predicted missions serve as language instructions for a VLA
policy, reducing human teleoperation time by 34.6\% relative to
GPT-4o and 39.8\% relative to GPT-5.6 Luna. 
\end{itemize}
VI~AMO, the annotation pipeline, and the VIM weights will be released at \href{https://see-want.github.io/}{https://see-want.github.io/}.

\section{RELATED WORK}
\label{sec:related}

\subsection{Operator Intent Inference}

Early shared-autonomy methods~\cite{Dragan-2012-7532,dragan2013policy,javdani2015hindsight,jain2019probabilistic,admoni2016predicting}
estimate user goals from control inputs or gaze to guide assistance.
Recent approaches use different observations and intent representations.
VOSA~\cite{belsare2025zeroshotuserintentrecognition} constructs object-goal
candidates from visual perception, while AHEAD~\cite{kim2026ahead} predicts
grasp objects and placement targets from tracked hand and head motion with
scene context.
CASPER~\cite{liu2025casper} infers a skill--object pair from visual
teleoperation history for user-confirmed skill execution.
Adaptor~\cite{liu2026adaptoradvancingassistiveteleoperation} learns latent
intention representations from teleoperation trajectories to condition action
generation. These representations support target selection, skill execution, or action
generation, but these functions do not by themselves require separate
descriptions of the current manipulation and its mission.
That distinction matters for continuation because the same action can serve
different goals.
We therefore study joint prediction of the ongoing atomic action and the
immediate mission it serves, making action recognition and goal inference
separately evaluable.

\subsection{Hierarchical Intent Supervision and Benchmarks}

Supervision for hierarchical intent prediction links observed actions to the
goals they serve.
In human activity understanding,
Ego4D Goal-Step~\cite{song2023ego4dgoalstep} annotates goals, steps, and
substeps, while AntGPT~\cite{zhao2024antgpt} infers goals from recognized
action histories to anticipate future actions.
For robots, BridgeData V2~\cite{walke2023bridgedata},
DROID~\cite{khazatsky2024droid}, and Open X-Embodiment~\cite{openx2023}
provide diverse manipulation demonstrations.
RH20T-P~\cite{chen2024rh20tp} adds manually segmented primitive skills, and
RoboVQA~\cite{sermanet2024robovqa} supports multi-horizon video reasoning.
Embodied chain-of-thought reasoning
(ECoT)~\cite{zawalski2024ecot} uses intermediate reasoning supervision for
instruction-conditioned control, while IntentVLA~\cite{lian2026intentvla}
uses visual history for instruction-conditioned action generation and
introduces AliasBench to evaluate ambiguous continuations. Building on these resources, we ask whether an ongoing robot action and its
mission can be jointly inferred before action completion, without the task
instruction.
Answering this question requires paired semantic targets aligned with the
observations available at each prediction time.
VI~AMO therefore associates temporally bounded atomic actions with their
immediate missions and evaluates both predictions on episode-disjoint data
at multiple stages of action progress.

\begin{figure*}[t]
 \centering
 \includegraphics[width=0.84\textwidth]{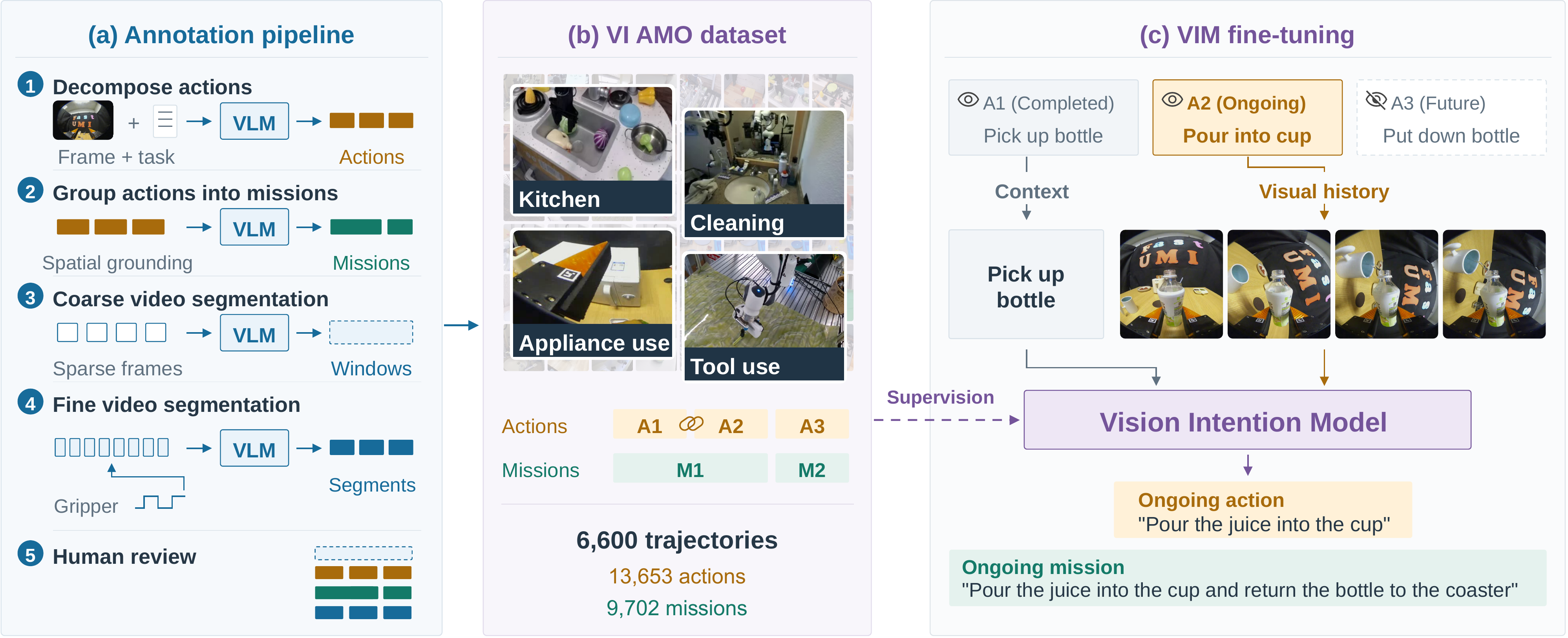} \caption{\textbf{From robot demonstrations to hierarchical intent.} (a)~Annotation pipeline. A VLM decomposes each episode instruction into ordered atomic actions and groups consecutive actions into the mission they serve; coarse-to-fine video segmentation localizes every action in time, with gripper signals guiding fine boundaries where available, followed by human review.
(b)~ Overview of our proposed VI AMO dataset. VI AMO pairs each atomic action with its mission across 6,600 episodes from eleven robot sources (13,653 actions, 9,702 missions), providing hierarchical supervision for intent prediction.
(c)~VIM is fine-tuned on VI AMO to generate the ongoing action and the ongoing mission from a short visual history and the completed-action context.}
 \label{fig:overview}
\end{figure*}

\section{PROBLEM FORMULATION}
\label{sec:problem}

\subsection{Hierarchical Intent Prediction}
\label{sec:io}
An \emph{atomic action} is an individual manipulation step, such as ``pick up the apple'' or ``place the apple in the basket''. A \emph{mission} is a sequence of consecutive atomic actions serving a single goal, such as ``put the apple in the basket''; the two form a hierarchy of goals and steps. Whether consecutive actions belong to the same mission is a commonsense judgment based on the scene: picking up an apple next to an empty basket naturally continues into placing it there, whereas an action without an evident continuation forms a mission on its own. At decision time $t$, the \emph{ongoing action} $\OA_t$ and \emph{ongoing mission} $\OM_t$ denote the action and mission the operator is currently executing.

Given these definitions, let $\hist_t$ denote $n$ time-ordered visual observations collected up to time $t$, and $\ctx_t$ an execution context summarizing behavior completed before $t$. An intention model $f_\phi$ predicts
\begin{equation}
 (\widehat{\OA}_t,\widehat{\OM}_t)=f_\phi(\hist_t,\ctx_t),
 \label{eq:intent}
\end{equation}
where both outputs are free-form natural-language descriptions generated without a predefined set of candidates. Both levels are needed because the same action can serve different missions: grasping an apple may precede placing it in a basket or handing it to a person.

\subsection{Application: Shared Autonomy}
\label{sec:handoff}
In shared autonomy, a robot autonomously continues a task initiated through teleoperation. We realize this with a language-conditioned VLA policy
\begin{equation}
 \mathbf{a}_t=\pi(\mathbf{o}_t,\mathbf{s}_t,\ell),
 \label{eq:vla}
\end{equation}
with action chunk $\mathbf{a}_t$, observations $\mathbf{o}_t$, robot state $\mathbf{s}_t$, and language instruction $\ell$. Since the operator provides no instruction, the intention model supplies it: at the switch time $t_h$, the robot executes
\begin{equation}
 \mathbf{a}_t=\pi\big(\mathbf{o}_t,\mathbf{s}_t,\widehat{\OM}_{t_h}\big),
 \quad (\,\cdot\,,\widehat{\OM}_{t_h})=f_\phi(\hist_{t_h},\ctx_{t_h}),
 \label{eq:sa}
\end{equation}
for all $t\geq t_h$, i.e., the predicted mission from~\eqref{eq:intent} serves as the instruction in~\eqref{eq:vla}. We assume a single switch and use only the teleoperation history up to $t_h$; the repeated switches in our real-world experiments apply the same procedure at each switch.

\section{METHOD}
\label{sec:method}
Our method centers on learning hierarchical intent from partial visual observations. We first introduce VI AMO, which pairs temporally localized atomic actions with the missions they serve. These annotations support both supervised learning and a benchmark for predicting ongoing action (OA) and ongoing mission (OM). We then train the Vision-Intention Model (VIM) to generate both descriptions from visual history and the completed-action context (Fig.~\ref{fig:overview}). Finally, we use the predicted mission as a language interface to a VLA policy, demonstrating shared autonomy as a downstream application.

\subsection{VI AMO: Hierarchical Action--Mission Annotations}
\label{sec:data}
Learning to infer intent requires supervision that distinguishes the observed manipulation from the goal it serves. VI AMO organizes robot demonstrations into temporally bounded atomic actions and their associated missions. Each atomic interval is paired with an OA description and an OM target.

As illustrated in Fig.~\ref{fig:overview}(a), our annotation pipeline consists of five stages.
(1) \textit{Action decomposition.} Given its strong spatial reasoning capabilities, we use GPT-5.6 Sol to identify task-relevant objects and propose ordered atomic-action descriptions from the episode instruction and first-frame images.
(2) \textit{Mission grouping.} We then prompt GPT-5.6 Sol with the task instruction, first-frame images, and ordered atomic actions to assess adjacent action pairs. Actions with an unambiguous continuation toward a shared goal are grouped into a mission; otherwise, we retain a mission boundary. Ungrouped actions then form single-action missions.
(3) \textit{Coarse temporal localization.} For episodes with multiple candidate atomic actions, Qwen3.7-Plus estimates a coarse temporal window for each atomic action from sparse frames spanning the video.
(4) \textit{Fine boundary refinement.} We select consecutive frames around each coarse boundary and nearby gripper closing or opening events, when available. Qwen3.7-Plus examines these frames to determine when the action is completed and refine the boundary. Each mission spans from the start of its first constituent action to the end of its last.
(5) \textit{Human review.} Human reviewers check and correct action descriptions, mission groupings, and temporal boundaries in around 50\% of the total episodes.

In total, this annotation pipeline yields 13,653 atomic-action segments and 9,702 mission segments across 6,600 episodes, as summarized in Table~\ref{tab:data}.

\begin{table}[!t]
\centering
\begin{threeparttable}
\caption{\textbf{VI AMO source composition with annotation summary.}}
\label{tab:data}\footnotesize\setlength{\tabcolsep}{2.5pt}
\begin{tabular*}{\columnwidth}{@{\extracolsep{\fill}}lrrrrr@{}}\toprule
Source & Train & Val & Test & Actions & Missions\\\midrule
Berkeley Fanuc~\cite{zhu2023fanuc} & 23 & 2 & 2 & 48 & 32 \\
Bridge V2~\cite{walke2023bridgedata} & 2,574 & 259 & 258 & 4,950 & 3,445 \\
CMU Play Fusion~\cite{chen2023playfusion} & 186 & 19 & 19 & 725 & 448 \\
DROID~\cite{khazatsky2024droid} & 1,702 & 169 & 170 & 6,372 & 4,452 \\
FastUMI-100K~\cite{liu2025fastumi100k} & 22 & 2 & 2 & 100 & 52 \\
Fractal (RT-1)~\cite{brohan2023rt1} & 483 & 48 & 48 & 656 & 579 \\
Jaco Play~\cite{dass2023jacoplay} & 56 & 6 & 6 & 68 & 68 \\
RH20T cfg 5~\cite{fang2024rh20t} & 79 & 8 & 8 & 224 & 143 \\
HYDRA~\cite{belkhale2023hydra} & 3 & 0 & 0 & 19 & 9 \\
TACO Play~\cite{rosete2023tacorl} & 336 & 34 & 33 & 403 & 403 \\
UT Austin MUTEX~\cite{shah2023mutex} & 36 & 3 & 4 & 88 & 71 \\
\midrule Total & 5,500 & 550 & 550 & 13,653 & 9,702 \\
\bottomrule
\end{tabular*}
\begin{tablenotes}[flushleft]\footnotesize
\item[] Episodes per split and annotated actions and missions per source.
\end{tablenotes}
\end{threeparttable}
\end{table}

\subsection{Hierarchical Intent Learning}
\label{sec:decoding}

To train VIM to recognize intent from visual history at different stages of an action, we construct six observation sequences for each atomic interval. Four use sampling windows beginning at action onset, with different lengths up to the full action duration. Within each partial window, we divide time into $n$ bins and randomly select one frame per bin. A fifth sequence samples a shorter local window at a random position within the action; and the sixth uses randomly spaced frames to provide additional variation in observation timing. We construct the visual history $\hist_t$ from one or more available camera views, sampling $n$ observations per view in chronological order as input to the VLM. 

Execution context $\ctx_t$ summarizes completed behavior in text form. In our training configuration, it lists completed actions within the active mission in chronological order. If the list is empty, the context is set to ``Robot has no previous actions.''

We instantiate VIM with Qwen3.5-27B~\cite{qwen2026qwen35} for its strong vision--language understanding capabilities and mature training ecosystem. The prompt defines the two intent levels and requests a structured response:
\begin{quote}\small\ttfamily
Ongoing action: \textless atomic action\textgreater\\
Ongoing mission: \textless mission\textgreater
\end{quote}
Both descriptions are generated in a single autoregressive pass. This exposes the model's interpretation of the current manipulation alongside the goal proposed for continuation. The output is free-form language, with no candidate-label bank supplied during generation.

\label{sec:training}
We adapt the backbone using low-rank adaptation (LoRA)~\cite{hu2022lora} in the vision and language attention and MLP modules due to memory efficiency. For a minibatch of $B$ examples, let $y_i=(y_{i,1},\ldots,y_{i,T_i})$ denote the supervised tokens of the formatted OA--OM response. We minimize the response-only autoregressive cross-entropy
\begin{equation}
 \mathcal{L}_{\mathrm{VIM}}(\phi)
 =-\frac{\displaystyle\sum_{i=1}^{B}\sum_{j=1}^{T_i}
 \log p_\phi(y_{i,j}\mid\hist_i,\ctx_i,y_{i,:j})}
 {\displaystyle\sum_{i=1}^{B}T_i}.
 \label{eq:vi_loss}
\end{equation}
Here $y_{i,:j}$ contains the preceding ground-truth response tokens. Template tokens are excluded from the loss, so supervision applies only to the predicted OA and OM. Only the low-rank parameters are updated; the pretrained weights remain frozen. Detailed optimization settings are provided in Sec.~\ref{sec:dataset_setup}.

\subsection{VLA Fine-Tuning for Shared Autonomy}
\label{sec:method_handoff}

In our downstream task, we integrate VIM with a VLA policy to
autonomously continue manipulation initiated through teleoperation.
We fine-tune a $\pio$ policy~\cite{intelligence2025pi05visionlanguageactionmodelopenworld}
on instruction-conditioned robot demonstrations using conditional flow-matching objective~\cite{physicalintelligence2026openpi}.
The policy predicts action chunks conditioned on current images,
robot state, and a language instruction,
$\mathbf{c}_t=(\mathbf{o}_t,\mathbf{s}_t,\ell)$.
For a demonstration action chunk $\mathbf{A}_t$, we construct
$\mathbf{A}_t^\tau=(1-\tau)\mathbf{A}_t+\tau\boldsymbol{\epsilon}$,
where $\boldsymbol{\epsilon}\sim\mathcal{N}(\mathbf{0},\mathbf{I})$
and the noise level $\tau$ follows OpenPI's sampling distribution.
The training objective is
\begin{equation}
\mathcal{L}_{\mathrm{VLA}}(\theta)
=
\mathbb{E}\!\left[
\left\|
v_\theta(\mathbf{A}_t^\tau,\tau,\mathbf{c}_t)
-
(\boldsymbol{\epsilon}-\mathbf{A}_t)
\right\|_F^2
\right],
\label{eq:vla_loss}
\end{equation}
where $v_\theta$ is the vector field predicted by the action expert
and $\ell$ is the demonstration instruction during training.

At deployment, VIM predicts the ongoing action and mission from
the teleoperation history. The operator reviews the predicted OM
and, if it matches their goal, approves it as the VLA's language
instruction for autonomous continuation.

\section{EXPERIMENTS}
\label{sec:exp}

\begin{table*}[!t]
\centering
\begin{threeparttable}
\caption{\textbf{OA and OM prediction on the VI~AMO test set.}}
\label{tab:main}
\footnotesize\setlength{\tabcolsep}{2.5pt}
\begin{tabular*}{\textwidth}{@{\extracolsep{\fill}}llrrrrrrrr@{}}\toprule
& & \multicolumn{4}{c}{Ongoing action} & \multicolumn{4}{c}{Ongoing mission}\\
\cmidrule(lr){3-6}\cmidrule(lr){7-10}
Dataset & Model & \metricup{F1} & \metricup{SBERT} & \metricup{SigLIP} & \metricup{Top-10} & \metricup{F1} & \metricup{SBERT} & \metricup{SigLIP} & \metricup{Top-10}\\\midrule
VI AMO test & GPT-5.6 Luna & 0.484 & \underline{0.581} & 0.741 & \underline{39.14} & 0.399 & 0.554 & 0.716 & \underline{39.33} \\
VI AMO test & GPT-4o & 0.463 & 0.563 & 0.733 & 35.32 & 0.330 & 0.470 & 0.685 & 29.91 \\
VI AMO test & Qwen Base (27B) & \underline{0.493} & \underline{0.581} & \underline{0.744} & 36.20 & \underline{0.424} & \underline{0.564} & \underline{0.722} & 37.27 \\
VI AMO test & Qwen Base (9B) & 0.464 & 0.529 & 0.718 & 31.43 & 0.368 & 0.496 & 0.684 & 28.32 \\
VI AMO test & RynnBrain~1.1-9B & 0.365 & 0.505 & 0.721 & 34.80 & 0.381 & 0.499 & 0.712 & 35.49 \\
VI AMO test & RoboBrain~2.5-8B & 0.236 & 0.316 & 0.631 & 18.87 & 0.205 & 0.315 & 0.577 & 19.29 \\
\midrule
VI AMO test & VIM (Ours) & \textbf{0.784} & \textbf{0.825} & \textbf{0.879} & \textbf{76.59} & \textbf{0.756} & \textbf{0.821} & \textbf{0.872} & \textbf{78.77} \\
\bottomrule
\end{tabular*}
\begin{tablenotes}[flushleft]\footnotesize
\item[] $\uparrow$: higher is better; Top-10 is in \%. \textbf{Bold}/\underline{underlined}: best/second-best within each dataset.
\end{tablenotes}
\end{threeparttable}
\end{table*}

We first evaluate OA and OM prediction on the VI AMO test set, evaluating both token-level accuracy and cosine similarity between predicted and ground-truth text embeddings. We then conduct ablation studies to examine the effects of camera views, textual context, visual history.
Finally, to demonstrate VIM's utility for downstream intent prediction tasks, we evaluate VIM on a physical Franka Emika Panda robot arm in five shared-autonomy manipulation tasks and measure the amount of manual teleoperation required during task execution.

\subsection{VI AMO Experimental Setup}
\label{sec:dataset_setup}
\label{sec:setup}

We fine-tune Qwen3.5-27B for one epoch in BF16 on 11,481 training actions, with $n=4$ observations per available camera and six temporal sampling plans per action (Sec.~\ref{sec:decoding}), yielding 68,886 training examples. We use LoRA with rank $r=16$, $\alpha=16$, and zero dropout. The effective batch size is 16 (four examples and four accumulation steps), and the sequence limit is 2,048 tokens. The learning rate is $10^{-4}$, with 3\% warm-up and weight decay 0.001. Training takes approximately 13.44 GPU-hours on a single NVIDIA A100 80GB GPU.

We evaluate VIM on the VI AMO test set. Its 550 episodes are disjoint from training and contain 1,091 atomic actions and 781 mission instances. Table~\ref{tab:data} summarizes the dataset composition. To assess intent prediction at different stages of manipulation, we evaluate at nine relative times: 0\%, 5\%, 15\%, 25\%, 40\%, 55\%, 70\%, 85\%, and 95\% of each atomic action's duration. These evaluation times cover the action from its onset to near completion; for example, 25\% means that one quarter of the action has elapsed. At each time, four frames are sampled uniformly from action onset to that time. At 0\%, the initial frame is padded four times.

We compare VIM with six baselines: the untuned Qwen3.5-27B backbone and the smaller Qwen3.5-9B~\cite{qwen2026qwen35}; two proprietary models, GPT-5.6 Luna (\texttt{gpt-5.6-luna}) and GPT-4o (\texttt{gpt-4o-2024-11-20}); and two open embodied foundation models post-trained for robot reasoning, RynnBrain~1.1-9B~\cite{li2026rynnbrain11} and RoboBrain~2.5-8B~\cite{tan2026robobrain25}. Front and wrist image pairs are used where available, with single-view fallback otherwise. Qwen uses deterministic decoding with thinking disabled and at most 80 new tokens.

\subsection{Evaluation Metrics}
\label{sec:metrics}

To comprehensively evaluate intent predictions, we measure both token-level agreement and embedding-level semantic similarity. We evaluate OA and OM separately because correctly identifying the immediate action does not necessarily imply understanding the broader mission it serves. \textit{Token F1} measures F1 score between the prediction and ground truth intent tokens, allowing partial matches. \textit{Sentence-BERT} (\textit{SBERT}) measures semantic similarity through the cosine of their frozen \texttt{all-mpnet-base-v2} sentence embeddings~\cite{reimers2019sbert}. The \textit{SigLIP} score likewise compares the two texts using cosine similarity from the frozen SigLIP text encoder~\cite{zhai2023siglip}, motivated by its image-text alignment and relevance to downstream robot control. For text-to-label retrieval, we rank the unique reference labels of the VI AMO test set (726 action labels and 539 mission labels) by their SigLIP similarity to the prediction. \textit{Top-10} measures how often the correct label appears among the first 10 candidates. We use Top-10 because semantically similar candidate labels may rank above the ground-truth label even when the prediction captures the intended action or mission. For example, ``pick up the pot lid'' and ``pick up the silver pot lid'' may describe the same action in a given scene.

\subsection{VI AMO Intent Prediction}
\label{sec:main_results}
\textbf{Overall intent prediction accuracy.} Averaged over the nine shared evaluation times, VIM outperforms all six baselines on every reported OA and OM metric (Table~\ref{tab:main}). Relative to the strongest baseline on each metric, OA/OM SBERT increases from 0.581/0.564 to 0.825/0.821, SigLIP cosine from 0.744/0.722 to 0.879/0.872, and OA/OM Top-10 retrieval accuracy from
39.14\%/39.33\% to 76.59\%/78.77\%. Despite fine-tuning for robotic perception, RynnBrain and RoboBrain do not outperform general-purpose VLMs on hierarchical intent prediction, suggesting that inferring operator intent from motion remains a distinct challenge.

\begin{figure*}[!t]
\centering\includegraphics[width=0.84\textwidth]{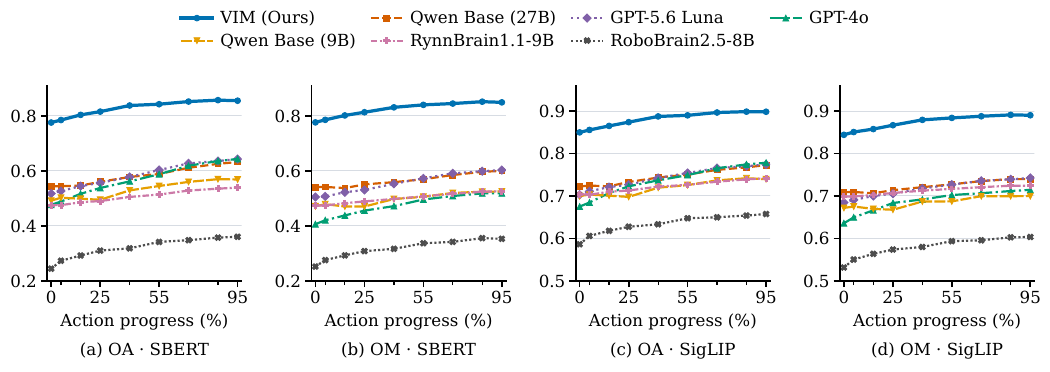}
\caption{\textbf{OA and OM prediction across action progress.} We evaluate ongoing action (OA) and ongoing mission (OM) predictions on the VI AMO test
set at nine observation cutoffs spanning 0–95\% of the current atomic action. Panels (a,b) report cosine similarity between predicted and ground-truth texts
using SBERT embeddings, while panels (c,d) use SigLIP text embeddings. Higher values indicate closer semantic agreement with the ground-truth intent. Curves connect the mean score at each observation cutoff, and all seven models use the same test examples, observation cutoffs, prompts, and execution context.}
\label{fig:prefix_curves}
\end{figure*}

\label{sec:progress}
\textbf{Intent recognition across action progress.} Fig.~\ref{fig:prefix_curves} shows that VIM achieves higher OA and OM similarity than all six baselines at all nine evaluation times under both SBERT and SigLIP. Its OA/OM SBERT rises from 0.776/0.776 at 0\% progress to 0.855/0.849 at 95\%, with smaller gains after 70\% progress. The baselines also improve with longer observation histories but remain below VIM near action completion. The consistent trends across both SBERT and SigLIP text encoders support recognizing the ongoing action and its mission from partial action observations and execution context, without waiting for the action to finish.

\begin{figure}[!tb]
\centering\includegraphics[width=0.99\columnwidth]{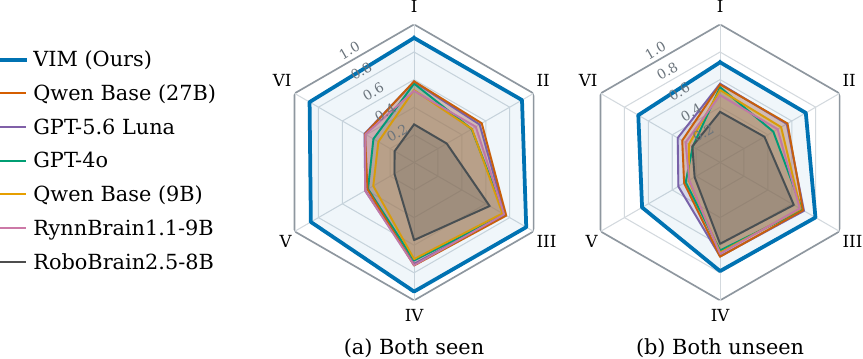}
\caption{\textbf{Seen versus unseen labels on the VI AMO test set.}
(a) Both OA and OM labels are seen during training; (b) both are unseen. Each polygon
represents one model; the blue outline denotes VIM. Axes run clockwise from the top: I: OA SBERT; II: OM SBERT; III: OA SigLIP; IV: OM SigLIP; V: OA Top-10; VI: OM Top-10.}
\label{fig:radar_seen_unseen}
\end{figure}

\label{sec:seen_unseen}
\textbf{Seen and unseen labels.} To evaluate the performance on previously observed and unseen labels, we split the 1,091 test actions by whether their normalized OA and OM label strings occur in the training annotations: 621 actions have both labels seen, 386 (35.4\%) have both labels unseen, and the remaining 84 are mixed (62 with only the OM label unseen and 22 with only the OA label unseen). Fig.~\ref{fig:radar_seen_unseen} compares all seven models on the two pure slices. On seen labels, VIM reaches OA/OM SBERT of 0.902/0.903, compared with 0.587/0.566 for the strongest baseline. On labels that never occur in training, VIM decreases to 0.726/0.717 SBERT but remains well above every baseline (at best 0.570/0.564).

\subsection{Ablation Study}
\label{sec:ablation}
We compare three variants of VIM: (1)~\emph{front-only} removes wrist images; (2)~\emph{no context} removes history text and samples from the current mission's onset to the action cutoff as the image-level context; and (3)~\emph{single @ 0\%} replaces the four-frame observation history with a single initial image pair, removing temporal motion cues from the input.
Table~\ref{tab:input_ablation_v3} summarizes the intent inference results for these variants. Removing the wrist view reduces OA/OM SBERT from 0.825/0.821 to 0.798/0.798, indicating that visual cues of gripper--object interaction provide complementary information for resolving fine-grained intent.
Removing the textual context reduces OA/OM SBERT to 0.767/0.788, and removing the visual history (a single observation at 0\%) reduces it to 0.793/0.792. The consistent decline in both OA and OM performance across the ablated variants supports the use of multi-view visual histories and completed-action context for hierarchical intent prediction.
\begin{table*}[!t]
\centering
\begin{threeparttable}
\caption{\textbf{Ablations on the VI AMO test set.}}
\label{tab:input_ablation_v3}
\footnotesize\setlength{\tabcolsep}{2.5pt}
\begin{tabular*}{\textwidth}{@{\extracolsep{\fill}}lrrrrrrrr@{}}\toprule
& \multicolumn{4}{c}{Ongoing action} & \multicolumn{4}{c}{Ongoing mission}\\
\cmidrule(lr){2-5}\cmidrule(lr){6-9}
Method & \metricup{F1} & \metricup{SBERT} & \metricup{SigLIP} & \metricup{Top-10} & \metricup{F1} & \metricup{SBERT} & \metricup{SigLIP} & \metricup{Top-10}\\\midrule
Front-only & \underline{0.753} & \underline{0.798} & \underline{0.861} & 70.13 & \underline{0.730} & \underline{0.798} & \underline{0.855} & \underline{73.75} \\
No context & 0.704 & 0.767 & 0.840 & 66.46 & 0.719 & 0.788 & 0.851 & 73.63 \\
Single @ 0\% & 0.750 & 0.793 & 0.858 & \underline{70.39} & 0.724 & 0.792 & 0.850 & 72.87 \\
\midrule
VIM (Ours) & \textbf{0.784} & \textbf{0.825} & \textbf{0.879} & \textbf{76.59} & \textbf{0.756} & \textbf{0.821} & \textbf{0.872} & \textbf{78.77} \\
\bottomrule
\end{tabular*}
\begin{tablenotes}[flushleft]\footnotesize
\item[] $\uparrow$: higher is better; Top-10 is in \%. \textbf{Bold}/\underline{underlined}: largest/second-largest displayed values per metric. 
\end{tablenotes}
\end{threeparttable}
\end{table*}

\begin{figure}[!t]
\setlength{\abovecaptionskip}{3pt}
\centering\includegraphics[width=1.0\linewidth]{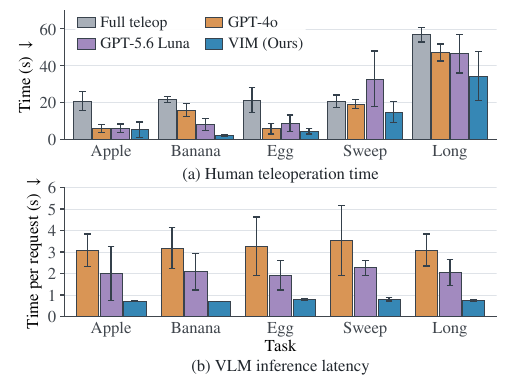}
\caption{\textbf{Real-world task quantitative results.} Bars show mean $\pm$ SD for (a) teleoperation time and (b) VLM latency.}
\label{fig:realworld_results}
\end{figure}

\begin{figure*}[!t]

\setlength{\abovecaptionskip}{3pt}
\centering\includegraphics[width=0.92\textwidth]{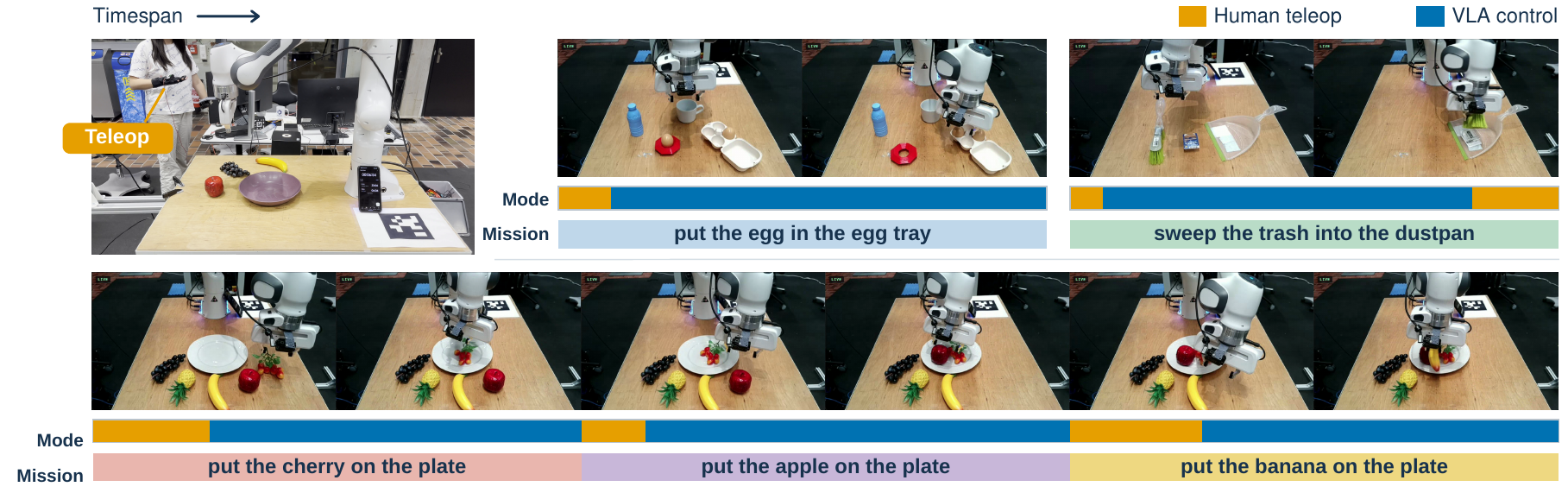}
\caption{\textbf{VIM-guided control handoffs across real-world tasks.} The top row shows, from left to right, the visualization of our shared autonomy setup,
egg placement, and trash sweeping task. The bottom row presents front-view keyframes of three successive missions in a multi-mission rollout: placing the cherry, apple, and banana on the plate. Mode bars indicate intervals of human teleoperation and VLA control, while mission bars identify the active mission. These examples illustrate the handoff process: VIM infers the ongoing mission from partial teleoperation, the operator confirms it, and the VLA continues execution.}
\label{fig:rollout}

\end{figure*}

\subsection{Real-World Experimental Setup}
\label{sec:teleoperation}
We evaluate VIM as the intent module in downstream shared-autonomy tasks. Our platform comprises a Franka arm with a parallel-jaw gripper, front/wrist OAK-D cameras, and a Rokoko Smartglove with Coil Pro tracking~\cite{rokokoStudio243} for teleoperation. Relative hand motion controls the end effector, and thumb--index pinch distance controls gripper opening.

VIM receives four uniformly sampled synchronized image pairs from the history starting from the latest user teleoperation takeover. 
The operator transfers control by confirming the inferred mission. The VLA then uses current observations and robot state to continue the approved task, issuing action commands at a nominal rate of 15\,Hz. When the operator resumes teleoperation, the glove mapping is re-anchored and robot commands again follow the operator. We deploy the VIM on an RTX~5090 GPU with TensorRT-Edge-LLM, quantized to NVFP4 to fit the available memory.

The five evaluation tasks, conducted in scenarios unseen during VIM fine-tuning, fall into four families, each probing a different aspect of intent understanding:
(1)~\emph{apple placement} and \emph{banana
placement} share a scene and differ only in the target fruit, so the mission
hinges on which object the operator approaches; (2)~\emph{egg placement} requires
the commonsense object--receptacle association (egg $\rightarrow$ tray);
(3)~\emph{trash sweeping} is mediated by a tool, so the goal must be read from
the implement the robot reaches for rather than from direct manipulation of the
target; and (4)~\emph{all fruits placement} demands repeated inference and
continuation across successive manipulations toward a compound goal. 

To fine-tune the VLA policy, we collect 130 teleoperated demonstrations for the all fruits placement task and 100 for each of the other tasks. We jointly fine-tune the policy on these demonstrations for 10,000 steps using four NVIDIA A100 GPUs with 80 GB of memory each.

We compare VIM-based shared autonomy with three baselines: GPT-4o and GPT-5.6 Luna as alternative intention models, and full teleoperation. We report total human teleoperation time and VLM latency. The results are based on five trials on each task, with mean $\pm$ sample standard deviation (SD) across trials.

\subsection{Real-World Experimental Results}
\label{sec:realworld_results}

The quantitative comparison among VIM, full teleoperation, and two baseline GPT models is illustrated in Fig.~\ref{fig:realworld_results}. Fig.~\ref{fig:rollout} provides qualitative examples of human-to-VLA handoffs across the egg-placement, sweeping, and long-horizon fruit-placement tasks.  Fig.~\ref{fig:realworld_results}(a) compares teleoperation time across the five task conditions. VIM requires less teleoperation than either GPT-based setup on every task, reducing the mean by 34.6\% relative to GPT-4o and 39.8\% relative to Luna.
Notably, in the trash-sweeping task, neither GPT model predicts a usable intent and the task therefore remains fully teleoperated.
Fig.~\ref{fig:realworld_results}(b) reports per-request mean and SD latency. SD describes variation among requests. VIM's mean latency ranges from 0.70 to 0.81\,s across tasks, compared with 3.09--3.53\,s for GPT-4o and 1.92--2.27\,s for Luna, achieving 76.9\% and 63.8\% latency reduction. Since many manipulation actions are completed over only a few seconds
and involve relatively short motions, delays above two seconds can cause the inferred intent to lag behind the robot's current state, making timely handoff difficult. 

\section{CONCLUSION AND LIMITATIONS}
\label{sec:conclusion}
We asked whether a VLM can decode an operator's intent from visual teleoperation history alone, and argued that intent should be represented hierarchically as the ongoing action and the mission it serves. To this end we introduced VI~AMO, a dataset of 6,600 episodes with temporally aligned action--mission annotations, and VIM, a VLM fine-tuned on it. On VI~AMO, VIM outperforms GPT-4o, GPT-5.6 Luna, two embodied foundation models, and its untuned backbone on every action and mission metric and at every stage of action progress (0.825/0.821 versus at most 0.581/0.564 SBERT for action/mission), with a smaller but consistent margin on label strings absent from training (0.726/0.717 versus 0.570/0.564), and it transfers to a real-world environment outside the dataset. In a five-task real-world study, its predicted mission let a VLA policy continue the task and reduced the operator's teleoperation time by over one-third compared with either GPT-based baseline, at 0.7\,s per prediction on a consumer GPU.

Although our framework improves hierarchical intent prediction and reduces manual teleoperation in downstream tasks, it still makes prediction errors and lacks calibrated confidence estimates. Reliably determining when its predictions can be trusted therefore remains an open challenge. Future work will expand VI AMO with more diverse demonstrations, potentially including synthetic data, to improve prediction accuracy and generalization ability. In addition, evaluation on the VI AMO dataset assumes predefined action boundaries and does not assess online boundary detection during live teleoperation. In our downstream real-world experiments, the VLA policy executes actions more slowly than a skilled operator, so reducing manual control does not always correspond to shorter task completion times. Future work will explore more lightweight action-generation architectures to reduce computational overhead~\cite{xie2026turbovla} and extend the framework to navigation and mobile manipulation.

\section*{\small ACKNOWLEDGEMENTS}

{\small
This research was supported by the Swiss National Science Foundation through the National Centre of Competence in Automation (NCCR automation), by an ETH RobotX research grant, and by grants from NVIDIA and utilized NVIDIA Jetson Thor.
\par}

\bibliographystyle{IEEEtran}
\bibliography{references-merged}

@preamble{ "\providecommand{\viaendwebtitle}[1]{}" }

@IEEEtranBSTCTL{VIA:referenceStyle,
  CTLuse_forced_etal       = {yes},
  CTLmax_names_forced_etal = {6},
  CTLnames_show_etal       = {1},
  CTLdash_repeated_names  = {no},
  CTLname_format_string   = {{f.~}{vv~}{ll}{ jj}},
  CTLuse_url             = {yes},
  CTLuse_alt_spacing     = {yes},
  CTLalt_stretch_factor  = {1.5}
}

@inproceedings{ahn2022saycan,
  author               = {Ichter, Brian and Brohan, Anthony and Chebotar, Yevgen and Finn, Chelsea and Hausman, Karol and Herzog, Alexander and Ho, Daniel and Ibarz, Julian and Irpan, Alex and Jang, Eric and Julian, Ryan and Kalashnikov, Dmitry and Levine, Sergey and Lu, Yao and Parada, Carolina and Rao, Kanishka and Sermanet, Pierre and Toshev, Alexander T and Vanhoucke, Vincent and Xia, Fei and Xiao, Ted and Xu, Peng and Yan, Mengyuan and Brown, Noah and Ahn, Michael and Cortes, Omar and Sievers, Nicolas and Tan, Clayton and Xu, Sichun and Reyes, Diego and Rettinghouse, Jarek and Quiambao, Jornell and Pastor, Peter and Luu, Linda and Lee, Kuang-Huei and Kuang, Yuheng and Jesmonth, Sally and Joshi, Nikhil J. and Jeffrey, Kyle and Ruano, Rosario Jauregui and Hsu, Jasmine and Gopalakrishnan, Keerthana and David, Byron and Zeng, Andy and Fu, Chuyuan Kelly},
  title                = {Do As {I} Can, Not As {I} Say: Grounding Language in Robotic Affordances},
  booktitle            = {Proc. 6th Conf. Robot Learn. (CoRL)},
  volume               = {205},
  pages                = {287--318},
  year                 = {2023},
  sourcebooktitle      = {Proceedings of The 6th Conference on Robot Learning},
  sourcepublisher      = {PMLR},
  sourceseries         = {Proceedings of Machine Learning Research},
  sourceurl            = {https://proceedings.mlr.press/v205/ichter23a.html},
}

@inproceedings{liang2023code,
  author               = {Liang, Jacky and Huang, Wenlong and Xia, Fei and Xu, Peng and Hausman, Karol and Ichter, Brian and Florence, Pete and Zeng, Andy},
  title                = {Code as Policies: Language Model Programs for Embodied Control},
  booktitle            = {Proc. IEEE Int. Conf. Robot. Automat. (ICRA)},
  pages                = {9493--9500},
  year                 = {2023},
  note                 = {doi: {\urlstyle{same}\url{10.1109/ICRA48891.2023.10160591}}},
  doi                  = {10.1109/ICRA48891.2023.10160591},
  sourcebooktitle      = {2023 IEEE International Conference on Robotics and Automation (ICRA)},
  sourcepublisher      = {IEEE},
  sourceurl            = {https://doi.org/10.1109/ICRA48891.2023.10160591},
}

@inproceedings{huang2022inner,
  author               = {Huang, Wenlong and Xia, Fei and Xiao, Ted and Chan, Harris and Liang, Jacky and Florence, Pete and Zeng, Andy and Tompson, Jonathan and Mordatch, Igor and Chebotar, Yevgen and Sermanet, Pierre and Jackson, Tomas and Brown, Noah and Luu, Linda and Levine, Sergey and Hausman, Karol and Ichter, Brian},
  title                = {Inner Monologue: Embodied Reasoning through Planning with Language Models},
  booktitle            = {Proc. 6th Conf. Robot Learn. (CoRL)},
  volume               = {205},
  pages                = {1769--1782},
  year                 = {2023},
  sourcebooktitle      = {Proceedings of The 6th Conference on Robot Learning},
  sourcepublisher      = {PMLR},
  sourceseries         = {Proceedings of Machine Learning Research},
  sourceurl            = {https://proceedings.mlr.press/v205/huang23c.html},
}

@inproceedings{zitkovich2023rt2,
  author               = {Zitkovich, Brianna and Yu, Tianhe and Xu, Sichun and Xu, Peng and Xiao, Ted and Xia, Fei and Wu, Jialin and Wohlhart, Paul and Welker, Stefan and Wahid, Ayzaan and Vuong, Quan and Vanhoucke, Vincent and Tran, Huong and Soricut, Radu and Singh, Anikait and Singh, Jaspiar and Sermanet, Pierre and Sanketi, Pannag R. and Salazar, Grecia and Ryoo, Michael S. and Reymann, Krista and Rao, Kanishka and Pertsch, Karl and Mordatch, Igor and Michalewski, Henryk and Lu, Yao and Levine, Sergey and Lee, Lisa and Lee, Tsang-Wei Edward and Leal, Isabel and Kuang, Yuheng and Kalashnikov, Dmitry and Julian, Ryan and Joshi, Nikhil J. and Irpan, Alex and Ichter, Brian and Hsu, Jasmine and Herzog, Alexander and Hausman, Karol and Gopalakrishnan, Keerthana and Fu, Chuyuan and Florence, Pete and Finn, Chelsea and Dubey, Kumar Avinava and Driess, Danny and Ding, Tianli and Choromanski, Krzysztof Marcin and Chen, Xi and Chebotar, Yevgen and Carbajal, Justice and Brown, Noah and Brohan, Anthony and Arenas, Montserrat Gonzalez and Han, Kehang},
  title                = {{RT-2}: Vision-Language-Action Models Transfer Web Knowledge to Robotic Control},
  booktitle            = {Proc. 7th Conf. Robot Learn. (CoRL)},
  volume               = {229},
  pages                = {2165--2183},
  year                 = {2023},
  sourcebooktitle      = {Proceedings of The 7th Conference on Robot Learning},
  sourcepublisher      = {PMLR},
  sourceseries         = {Proceedings of Machine Learning Research},
  sourceurl            = {https://proceedings.mlr.press/v229/zitkovich23a.html},
}

@inproceedings{kim2024openvlaopensourcevisionlanguageactionmodel,
  author               = {Kim, Moo Jin and Pertsch, Karl and Karamcheti, Siddharth and Xiao, Ted and Balakrishna, Ashwin and Nair, Suraj and Rafailov, Rafael and Foster, Ethan P and Sanketi, Pannag R and Vuong, Quan and Kollar, Thomas and Burchfiel, Benjamin and Tedrake, Russ and Sadigh, Dorsa and Levine, Sergey and Liang, Percy and Finn, Chelsea},
  title                = {{OpenVLA}: An Open-Source Vision-Language-Action Model},
  booktitle            = {Proc. 8th Conf. Robot Learn. (CoRL)},
  volume               = {270},
  pages                = {2679--2713},
  year                 = {2025},
  sourcebooktitle      = {Proceedings of The 8th Conference on Robot Learning},
  sourcepublisher      = {PMLR},
  sourceseries         = {Proceedings of Machine Learning Research},
  sourceurl            = {https://proceedings.mlr.press/v270/kim25c.html},
}

@inproceedings{black2026pi0visionlanguageactionflowmodel,
  author               = {Kevin Black and Noah Brown and Danny Driess and Adnan Esmail and Michael Robert Equi and Chelsea Finn and Niccolo Fusai and Lachy Groom and Karol Hausman and Brian Ichter and Szymon Jakubczak and Tim Jones and Liyiming Ke and Sergey Levine and Adrian Li-Bell and Mohith Mothukuri and Suraj Nair and Karl Pertsch and Lucy Xiaoyang Shi and Laura Smith and James Tanner and Quan Vuong and Anna Walling and Haohuan Wang and Ury Zhilinsky},
  title                = {{$\pi_0$}: A Vision-Language-Action Flow Model for General Robot Control},
  booktitle            = {Proc. Robot.: Sci. Syst. (RSS)},
  year                 = {2025},
  note                 = {doi: {\urlstyle{same}\url{10.15607/RSS.2025.XXI.010}}},
  doi                  = {10.15607/RSS.2025.XXI.010},
  sourcebooktitle      = {Proceedings of Robotics: Science and Systems},
  sourceurl            = {https://www.roboticsproceedings.org/rss21/p010.html},
}

@inproceedings{intelligence2025pi05visionlanguageactionmodelopenworld,
  author               = {Black, Kevin and Brown, Noah and Darpinian, James and Dhabalia, Karan and Driess, Danny and Esmail, Adnan and Equi, Michael Robert and Finn, Chelsea and Fusai, Niccolo and Galliker, Manuel Y. and Ghosh, Dibya and Groom, Lachy and Hausman, Karol and Ichter, Brian and Jakubczak, Szymon and Jones, Tim and Ke, Liyiming and LeBlanc, Devin and Levine, Sergey and Li-Bell, Adrian and Mothukuri, Mohith and Nair, Suraj and Pertsch, Karl and Ren, Allen Z. and Shi, Lucy Xiaoyang and Smith, Laura and Springenberg, Jost Tobias and Stachowicz, Kyle and Tanner, James and Vuong, Quan and Walke, Homer and Walling, Anna and Wang, Haohuan and Yu, Lili and Zhilinsky, Ury},
  title                = {{$\pi_{0.5}$}: A Vision-Language-Action Model with Open-World Generalization},
  booktitle            = {Proc. 9th Conf. Robot Learn. (CoRL)},
  volume               = {305},
  pages                = {17--40},
  year                 = {2025},
  sourcebooktitle      = {Proceedings of The 9th Conference on Robot Learning},
  sourcepublisher      = {PMLR},
  sourceseries         = {Proceedings of Machine Learning Research},
  sourceurl            = {https://proceedings.mlr.press/v305/black25a.html},
}

@inproceedings{belsare2025zeroshotuserintentrecognition,
  author               = {Belsare, Atharv and Karimi, Zohre and Mattson, Connor and Brown, Daniel S.},
  title                = {Toward Zero-Shot User Intent Recognition in Shared Autonomy},
  booktitle            = {Proc. ACM/IEEE Int. Conf. Human-Robot Interact. (HRI)},
  pages                = {222--231},
  year                 = {2025},
  note                 = {doi: {\urlstyle{same}\url{10.1109/HRI61500.2025.10973985}}},
  doi                  = {10.1109/HRI61500.2025.10973985},
  sourcebooktitle      = {Proceedings of the 2025 ACM/IEEE International Conference on Human-Robot Interaction (HRI)},
  sourceurl            = {https://doi.org/10.1109/HRI61500.2025.10973985},
}

@inproceedings{liu2025casper,
  author               = {Liu, Huihan and Shah, Rutav and Liu, Shuijing and Pittenger, Jack and Seo, Mingyo and Cui, Yuchen and Bisk, Yonatan and Mart{\'i}n-Mart{\'i}n, Roberto and Zhu, Yuke},
  title                = {{CASPER}: Inferring Diverse Intents for Assistive Teleoperation with Vision Language Models},
  booktitle            = {Proc. 9th Conf. Robot Learn. (CoRL)},
  volume               = {305},
  pages                = {2462--2483},
  year                 = {2025},
  sourcebooktitle      = {Proceedings of The 9th Conference on Robot Learning},
  sourcepublisher      = {PMLR},
  sourceseries         = {Proceedings of Machine Learning Research},
  sourceurl            = {https://proceedings.mlr.press/v305/liu25d.html},
}

@inproceedings{walke2023bridgedata,
  author               = {Walke, Homer Rich and Black, Kevin and Zhao, Tony Z. and Vuong, Quan and Zheng, Chongyi and Hansen-Estruch, Philippe and He, Andre Wang and Myers, Vivek and Kim, Moo Jin and Du, Max and Lee, Abraham and Fang, Kuan and Finn, Chelsea and Levine, Sergey},
  title                = {{BridgeData V2}: A Dataset for Robot Learning at Scale},
  booktitle            = {Proc. 7th Conf. Robot Learn. (CoRL)},
  volume               = {229},
  pages                = {1723--1736},
  year                 = {2023},
  sourcebooktitle      = {Proceedings of The 7th Conference on Robot Learning},
  sourcepublisher      = {PMLR},
  sourceseries         = {Proceedings of Machine Learning Research},
  sourceurl            = {https://proceedings.mlr.press/v229/walke23a.html},
}

@inproceedings{khazatsky2024droid,
  author               = {Alexander Khazatsky and Karl Pertsch and Suraj Nair and Ashwin Balakrishna and Sudeep Dasari and Siddharth Karamcheti and Soroush Nasiriany and Mohan Kumar Srirama and Lawrence Yunliang Chen and Kirsty Ellis and Peter David Fagan and Joey Hejna and Masha Itkina and Marion Lepert and Yecheng Jason Ma and Patrick Tree Miller and Jimmy Wu and Suneel Belkhale and Shivin Dass and Huy Ha and Arhan Jain and Abraham Lee and Youngwoon Lee and Marius Memmel and Sungjae Park and Ilija Radosavovic and Kaiyuan Wang and Albert Zhan and Kevin Black and Cheng Chi and Kyle Beltran Hatch and Shan Lin and Jingpei Lu and Jean Mercat and Abdul Rehman and Pannag R Sanketi and Archit Sharma and Cody Simpson and Quan Vuong and Homer Rich Walke and Blake Wulfe and Ted Xiao and Jonathan Heewon Yang and Arefeh Yavary and Tony Z. Zhao and Christopher Agia and Rohan Baijal and Mateo Guaman Castro and Daphne Chen and Qiuyu Chen and Trinity Chung and Jaimyn Drake and Ethan Paul Foster and Jensen Gao and David Antonio Herrera and Minho Heo and Kyle Hsu and Jiaheng Hu and Donovon Jackson and Charlotte Le and Yunshuang Li and Roy Lin and Zehan Ma and Abhiram Maddukuri and Suvir Mirchandani and Daniel Morton and Tony Nguyen and Abigail O'Neill and Rosario Scalise and Derick Seale and Victor Son and Stephen Tian and Emi Tran and Andrew E. Wang and Yilin Wu and Annie Xie and Jingyun Yang and Patrick Yin and Yunchu Zhang and Osbert Bastani and Glen Berseth and Jeannette Bohg and Ken Goldberg and Abhinav Gupta and Abhishek Gupta and Dinesh Jayaraman and Joseph J Lim and Jitendra Malik and Roberto Mart{\'i}n-Mart{\'i}n and Subramanian Ramamoorthy and Dorsa Sadigh and Shuran Song and Jiajun Wu and Michael C. Yip and Yuke Zhu and Thomas Kollar and Sergey Levine and Chelsea Finn},
  title                = {{DROID}: A Large-Scale In-The-Wild Robot Manipulation Dataset},
  booktitle            = {Proc. Robot.: Sci. Syst. (RSS)},
  year                 = {2024},
  note                 = {doi: {\urlstyle{same}\url{10.15607/RSS.2024.XX.120}}},
  doi                  = {10.15607/RSS.2024.XX.120},
  sourcebooktitle      = {Proceedings of Robotics: Science and Systems},
  sourceurl            = {https://www.roboticsproceedings.org/rss20/p120.html},
}

@inproceedings{openx2023,
  author               = {{Open X-Embodiment Collaboration} and Abby O'Neill and Abdul Rehman and Abhiram Maddukuri and Abhishek Gupta and Abhishek Padalkar and Abraham Lee and Acorn Pooley and Agrim Gupta and Ajay Mandlekar and Ajinkya Jain and Albert Tung and Alex Bewley and Alex Herzog and Alex Irpan and Alexander Khazatsky and Anant Rai and Anchit Gupta and Andrew Wang and Anikait Singh and Animesh Garg and Aniruddha Kembhavi and Annie Xie and Anthony Brohan and Antonin Raffin and Archit Sharma and Arefeh Yavary and Arhan Jain and Ashwin Balakrishna and Ayzaan Wahid and Ben Burgess-Limerick and Beomjoon Kim and Bernhard Sch{\"o}lkopf and Blake Wulfe and Brian Ichter and Cewu Lu and Charles Xu and Charlotte Le and Chelsea Finn and Chen Wang and Chenfeng Xu and Cheng Chi and Chenguang Huang and Christine Chan and Christopher Agia and Chuer Pan and Chuyuan Fu and Coline Devin and Danfei Xu and Daniel Morton and Danny Driess and Daphne Chen and Deepak Pathak and Dhruv Shah and Dieter B{\"u}chler and Dinesh Jayaraman and Dmitry Kalashnikov and Dorsa Sadigh and Edward Johns and Ethan Foster and Fangchen Liu and Federico Ceola and Fei Xia and Feiyu Zhao and Freek Stulp and Gaoyue Zhou and Sukhatme, Gaurav S. and Gautam Salhotra and Ge Yan and Gilbert Feng and Giulio Schiavi and Glen Berseth and Gregory Kahn and Guanzhi Wang and Hao Su and Hao-Shu Fang and Haochen Shi and Henghui Bao and Amor, Heni Ben and Christensen, Henrik I. and Hiroki Furuta and Homer Walke and Hongjie Fang and Huy Ha and Igor Mordatch and Ilija Radosavovic and Isabel Leal and Jacky Liang and Jad Abou-Chakra and Jaehyung Kim and Jaimyn Drake and Jan Peters and Jan Schneider and Jasmine Hsu and Jeannette Bohg and Jeffrey Bingham and Jeffrey Wu and Jensen Gao and Jiaheng Hu and Jiajun Wu and Jialin Wu and Jiankai Sun and Jianlan Luo and Jiayuan Gu and Jie Tan and Jihoon Oh and Jimmy Wu and Jingpei Lu and Jingyun Yang and Jitendra Malik and Jo{\~a}o Silv{\'e}rio and Joey Hejna and Jonathan Booher and Jonathan Tompson and Jonathan Yang and Jordi Salvador and Lim, Joseph J. and Junhyek Han and Kaiyuan Wang and Kanishka Rao and Karl Pertsch and Karol Hausman and Keegan Go and Keerthana Gopalakrishnan and Ken Goldberg and Kendra Byrne and Kenneth Oslund and Kento Kawaharazuka and Kevin Black and Kevin Lin and Kevin Zhang and Kiana Ehsani and Kiran Lekkala and Kirsty Ellis and Krishan Rana and Krishnan Srinivasan and Kuan Fang and Singh, Kunal Pratap and Kuo-Hao Zeng and Kyle Hatch and Kyle Hsu and Laurent Itti and Chen, Lawrence Yunliang and Lerrel Pinto and Li Fei-Fei and Liam Tan and Fan, Linxi Jim and Lionel Ott and Lisa Lee and Luca Weihs and Magnum Chen and Marion Lepert and Marius Memmel and Masayoshi Tomizuka and Masha Itkina and Castro, Mateo Guaman and Max Spero and Maximilian Du and Michael Ahn and Yip, Michael C. and Mingtong Zhang and Mingyu Ding and Minho Heo and Srirama, Mohan Kumar and Mohit Sharma and Kim, Moo Jin and Naoaki Kanazawa and Nicklas Hansen and Nicolas Heess and Joshi, Nikhil J. and Niko Suenderhauf and Ning Liu and Palo, Norman Di and Shafiullah, Nur Muhammad Mahi and Oier Mees and Oliver Kroemer and Osbert Bastani and Sanketi, Pannag R. and Miller, Patrick Tree and Patrick Yin and Paul Wohlhart and Peng Xu and Fagan, Peter David and Peter Mitrano and Pierre Sermanet and Pieter Abbeel and Priya Sundaresan and Qiuyu Chen and Quan Vuong and Rafael Rafailov and Ran Tian and Ria Doshi and Roberto Mart{\'i}n-Mart{\'i}n and Rohan Baijal and Rosario Scalise and Rose Hendrix and Roy Lin and Runjia Qian and Ruohan Zhang and Russell Mendonca and Rutav Shah and Ryan Hoque and Ryan Julian and Samuel Bustamante and Sean Kirmani and Sergey Levine and Shan Lin and Sherry Moore and Shikhar Bahl and Shivin Dass and Shubham Sonawani and Shuran Song and Sichun Xu and Siddhant Haldar and Siddharth Karamcheti and Simeon Adebola and Simon Guist and Soroush Nasiriany and Stefan Schaal and Stefan Welker and Stephen Tian and Subramanian Ramamoorthy and Sudeep Dasari and Suneel Belkhale and Sungjae Park and Suraj Nair and Suvir Mirchandani and Takayuki Osa and Tanmay Gupta and Tatsuya Harada and Tatsuya Matsushima and Ted Xiao and Thomas Kollar and Tianhe Yu and Tianli Ding and Todor Davchev and Zhao, Tony Z. and Travis Armstrong and Trevor Darrell and Trinity Chung and Vidhi Jain and Vincent Vanhoucke and Wei Zhan and Wenxuan Zhou and Wolfram Burgard and Xi Chen and Xiaolong Wang and Xinghao Zhu and Xinyang Geng and Xiyuan Liu and Xu Liangwei and Xuanlin Li and Yao Lu and Ma, Yecheng Jason and Yejin Kim and Yevgen Chebotar and Yifan Zhou and Yifeng Zhu and Yilin Wu and Ying Xu and Yixuan Wang and Yonatan Bisk and Yoonyoung Cho and Youngwoon Lee and Yuchen Cui and Yue Cao and Yueh-Hua Wu and Yujin Tang and Yuke Zhu and Yunchu Zhang and Yunfan Jiang and Yunshuang Li and Yunzhu Li and Yusuke Iwasawa and Yutaka Matsuo and Zehan Ma and Zhuo Xu and Cui, Zichen Jeff and Zichen Zhang and Zipeng Lin},
  title                = {{Open X-Embodiment}: Robotic Learning Datasets and {RT-X} Models},
  booktitle            = {Proc. IEEE Int. Conf. Robot. Automat. (ICRA)},
  pages                = {6892--6903},
  year                 = {2024},
  note                 = {doi: {\urlstyle{same}\url{10.1109/ICRA57147.2024.10611477}}},
  doi                  = {10.1109/ICRA57147.2024.10611477},
  sourcebooktitle      = {2024 IEEE International Conference on Robotics and Automation (ICRA)},
  sourcepublisher      = {IEEE},
  sourceurl            = {https://doi.org/10.1109/ICRA57147.2024.10611477},
}

@inproceedings{brohan2023rt1,
  author               = {Anthony Brohan and Noah Brown and Justice Carbajal and Yevgen Chebotar and Joseph Dabis and Chelsea Finn and Keerthana Gopalakrishnan and Karol Hausman and Alexander Herzog and Jasmine Hsu and Julian Ibarz and Brian Ichter and Alex Irpan and Tomas Jackson and Sally Jesmonth and Nikhil Joshi and Ryan Julian and Dmitry Kalashnikov and Yuheng Kuang and Isabel Leal and Kuang-Huei Lee and Sergey Levine and Yao Lu and Utsav Malla and Deeksha Manjunath and Igor Mordatch and Ofir Nachum and Carolina Parada and Jodilyn Peralta and Emily Perez and Karl Pertsch and Jornell Quiambao and Kanishka Rao and Michael S Ryoo and Grecia Salazar and Pannag R Sanketi and Kevin Sayed and Jaspiar Singh and Sumedh Sontakke and Austin Stone and Clayton Tan and Huong Tran and Vincent Vanhoucke and Steve Vega and Quan H Vuong and Fei Xia and Ted Xiao and Peng Xu and Sichun Xu and Tianhe Yu and Brianna Zitkovich},
  title                = {{RT-1}: Robotics Transformer for Real-World Control at Scale},
  booktitle            = {Proc. Robot.: Sci. Syst. (RSS)},
  year                 = {2023},
  note                 = {doi: {\urlstyle{same}\url{10.15607/RSS.2023.XIX.025}}},
  doi                  = {10.15607/RSS.2023.XIX.025},
  sourcebooktitle      = {Proceedings of Robotics: Science and Systems},
  sourceurl            = {https://www.roboticsproceedings.org/rss19/p025.html},
}

@inproceedings{zawalski2024ecot,
  author               = {Zawalski, Micha{\l} and Chen, William and Pertsch, Karl and Mees, Oier and Finn, Chelsea and Levine, Sergey},
  title                = {Robotic Control via Embodied Chain-of-Thought Reasoning},
  booktitle            = {Proc. 8th Conf. Robot Learn. (CoRL)},
  volume               = {270},
  pages                = {3157--3181},
  year                 = {2025},
  sourcebooktitle      = {Proceedings of The 8th Conference on Robot Learning},
  sourcepublisher      = {PMLR},
  sourceseries         = {Proceedings of Machine Learning Research},
  sourceurl            = {https://proceedings.mlr.press/v270/zawalski25a.html},
}

@misc{chen2024rh20tp,
  author               = {Chen, Zeren and Shi, Zhelun and Lu, Xiaoya and He, Lehan and Qian, Sucheng and Yin, Zhenfei and Ouyang, Wanli and Shao, Jing and Qiao, Yu and Lu, Cewu and Sheng, Lu},
  title                = {{RH20T-P}: A Primitive-Level Robotic Dataset Towards Composable Generalization Agents},
  year                 = {2024},
  note                 = {{arXiv:}2403.19622},
  doi                  = {10.48550/arXiv.2403.19622},
  eprint               = {2403.19622},
  archiveprefix        = {arXiv},
  primaryclass         = {cs.RO},
  sourcehowpublished   = {arXiv preprint arXiv:2403.19622},
  sourcenote           = {NeurIPS 2024 Workshop},
  sourceurl            = {https://arxiv.org/abs/2403.19622},
}

@article{9501975,
  author               = {Selvaggio, Mario and Cognetti, Marco and Nikolaidis, Stefanos and Ivaldi, Serena and Siciliano, Bruno},
  title                = {Autonomy in Physical Human-Robot Interaction: A Brief Survey},
  journal              = {IEEE Robot. Autom. Lett.},
  volume               = {6},
  number               = {4},
  pages                = {7989--7996},
  year                 = {2021},
  note                 = {doi: {\urlstyle{same}\url{10.1109/LRA.2021.3100603}}},
  doi                  = {10.1109/LRA.2021.3100603},
  sourcejournal        = {IEEE Robotics and Automation Letters},
  sourceurl            = {https://doi.org/10.1109/LRA.2021.3100603},
}

@article{losey2022learning,
  author               = {Losey, Dylan P. and Jeon, Hong Jun and Li, Mengxi and Srinivasan, Krishnan and Mandlekar, Ajay and Garg, Animesh and Bohg, Jeannette and Sadigh, Dorsa},
  title                = {Learning latent actions to control assistive robots},
  journal              = {Auton. Robots},
  volume               = {46},
  number               = {1},
  pages                = {115--147},
  year                 = {2022},
  note                 = {doi: {\urlstyle{same}\url{10.1007/s10514-021-10005-w}}},
  doi                  = {10.1007/s10514-021-10005-w},
  sourcejournal        = {Autonomous Robots},
  sourceurl            = {https://doi.org/10.1007/s10514-021-10005-w},
}

@inproceedings{Dragan-2012-7532,
  author               = {Dragan, Anca D. and Srinivasa, Siddhartha S.},
  title                = {Formalizing Assistive Teleoperation},
  booktitle            = {Proc. Robot.: Sci. Syst. (RSS)},
  year                 = {2012},
  note                 = {doi: {\urlstyle{same}\url{10.15607/RSS.2012.VIII.010}}},
  doi                  = {10.15607/RSS.2012.VIII.010},
  sourcebooktitle      = {Proceedings of Robotics: Science and Systems},
  sourceurl            = {https://www.roboticsproceedings.org/rss08/p10.html},
}

@article{dragan2013policy,
  author               = {Dragan, Anca D. and Srinivasa, Siddhartha S.},
  title                = {A Policy-Blending Formalism for Shared Control},
  journal              = {Int. J. Robot. Res.},
  volume               = {32},
  number               = {7},
  pages                = {790--805},
  year                 = {2013},
  note                 = {doi: {\urlstyle{same}\url{10.1177/0278364913490324}}},
  doi                  = {10.1177/0278364913490324},
  sourcejournal        = {The International Journal of Robotics Research},
  sourceurl            = {https://doi.org/10.1177/0278364913490324},
}

@inproceedings{javdani2015hindsight,
  author               = {Javdani, Shervin and Srinivasa, Siddhartha S. and Bagnell, J. Andrew},
  title                = {Shared Autonomy via Hindsight Optimization},
  booktitle            = {Proc. Robot.: Sci. Syst. (RSS)},
  year                 = {2015},
  note                 = {doi: {\urlstyle{same}\url{10.15607/RSS.2015.XI.032}}},
  doi                  = {10.15607/RSS.2015.XI.032},
  sourcebooktitle      = {Proceedings of Robotics: Science and Systems},
  sourceurl            = {https://www.roboticsproceedings.org/rss11/p32.html},
}

@article{jain2019probabilistic,
  author               = {Jain, Siddarth and Argall, Brenna},
  title                = {Probabilistic Human Intent Recognition for Shared Autonomy in Assistive Robotics},
  journal              = {ACM Trans. Human-Robot Interact.},
  volume               = {9},
  number               = {1},
  year                 = {2019},
  note                 = {{Art. no.} 2, doi: {\urlstyle{same}\url{10.1145/3359614}}},
  doi                  = {10.1145/3359614},
  articleno            = {2},
  numpages             = {23},
  sourcejournal        = {ACM Transactions on Human-Robot Interaction},
  sourceurl            = {https://doi.org/10.1145/3359614},
  sourcepages          = {2:1--2:23},
}

@inproceedings{admoni2016predicting,
  author               = {Admoni, Henny and Srinivasa, Siddhartha S.},
  title                = {Predicting User Intent Through Eye Gaze for Shared Autonomy},
  booktitle            = {Proc. AAAI Fall Symp. Shared Autonomy Res. Pract.},
  pages                = {298--303},
  year                 = {2016},
  sourcebooktitle      = {Proceedings of the AAAI Fall Symposium on Shared Autonomy in Research and Practice},
  sourceurl            = {https://publications.ri.cmu.edu/predicting-user-intent-through-eye-gaze-for-shared-autonomy},
}

@misc{kim2026ahead,
  author               = {Kim, Seok Joon and Lee, Junho and Spinola, Federica and Kwon, Taein and Moghaddam, Mohsen},
  title                = {{AHEAD}: Anticipatory Hand-Driven Teleoperation via Human Intent Prediction},
  year                 = {2026},
  note                 = {{arXiv:}2607.15172},
  eprint               = {2607.15172},
  archiveprefix        = {arXiv},
  primaryclass         = {cs.RO},
  sourceentrytype      = {inproceedings},
  sourcebooktitle      = {IEEE/RSJ International Conference on Intelligent Robots and Systems (IROS)},
  sourcenote           = {Accepted to IROS 2026; arXiv:2607.15172},
  sourceurl            = {https://arxiv.org/abs/2607.15172},
}

@misc{liu2026adaptoradvancingassistiveteleoperation,
  author               = {Liu, Yu and Yin, Yihang and Huang, Tianlv and Yan, Fei and Xu, Yuan and Hong, Weinan and Han, Wei and Cao, Yue and Chen, Xiangyu and Fan, Zipei and Song, Xuan},
  title                = {{Adaptor}: Advancing Assistive Teleoperation with Few-Shot Learning and Cross-Operator Generalization},
  year                 = {2026},
  note                 = {{arXiv:}2604.09462},
  eprint               = {2604.09462},
  archiveprefix        = {arXiv},
  sourceentrytype      = {inproceedings},
  sourcebooktitle      = {IEEE International Conference on Robotics and Automation (ICRA)},
  sourcenote           = {Accepted to ICRA 2026; arXiv:2604.09462},
  sourceurl            = {https://arxiv.org/abs/2604.09462},
}

@inproceedings{song2023ego4dgoalstep,
  author               = {Song, Yale and Byrne, Eugene and Nagarajan, Tushar and Wang, Huiyu and Martin, Miguel and Torresani, Lorenzo},
  title                = {{Ego4D Goal-Step}: Toward Hierarchical Understanding of Procedural Activities},
  booktitle            = {Adv. Neural Inf. Process. Syst.},
  volume               = {36},
  year                 = {2023},
  note                 = {doi: {\urlstyle{same}\url{10.52202/075280-1688}}},
  doi                  = {10.52202/075280-1688},
  sourcebooktitle      = {Advances in Neural Information Processing Systems},
  sourcenote           = {Datasets and Benchmarks Track},
  sourceurl            = {https://papers.neurips.cc/paper_files/paper/2023/hash/7a65606fa1a6849450550325832036e5-Abstract-Datasets_and_Benchmarks.html},
}

@inproceedings{zhao2024antgpt,
  author               = {Zhao, Qi and Wang, Shijie and Zhang, Ce and Fu, Changcheng and Do, Minh Quan and Agarwal, Nakul and Lee, Kwonjoon and Sun, Chen},
  title                = {{AntGPT}: Can Large Language Models Help Long-Term Action Anticipation from Videos?},
  booktitle            = {Proc. Int. Conf. Learn. Represent. (ICLR)},
  year                 = {2024},
  url                  = {https://arxiv.org/abs/2307.16368},
  sourcebooktitle      = {International Conference on Learning Representations (ICLR)},
}

@inproceedings{sermanet2024robovqa,
  author               = {Pierre Sermanet and Tianli Ding and Jeffrey Zhao and Fei Xia and Debidatta Dwibedi and Keerthana Gopalakrishnan and Christine Chan and Gabriel Dulac-Arnold and Sharath Maddineni and Nikhil J Joshi and Pete Florence and Wei Han and Robert Baruch and Yao Lu and Suvir Mirchandani and Peng Xu and Pannag Sanketi and Karol Hausman and Izhak Shafran and Brian Ichter and Yuan Cao},
  title                = {{RoboVQA}: Multimodal Long-Horizon Reasoning for Robotics},
  booktitle            = {Proc. IEEE Int. Conf. Robot. Automat. (ICRA)},
  pages                = {645--652},
  year                 = {2024},
  note                 = {doi: {\urlstyle{same}\url{10.1109/ICRA57147.2024.10610216}}},
  doi                  = {10.1109/ICRA57147.2024.10610216},
  sourcebooktitle      = {IEEE International Conference on Robotics and Automation (ICRA)},
  sourceurl            = {https://doi.org/10.1109/ICRA57147.2024.10610216},
}

@misc{lian2026intentvla,
  author               = {Lian, Shijie and Yu, Bin and Lin, Xiaopeng and Shen, Zhaolong and Yang, Laurence Tianruo and Jin, Yurun and Liu, Haishan and Wu, Changti and Yuan, Hang and Huang, Cong and Chen, Kai},
  title                = {{IntentVLA}: Short-Horizon Intent Modeling for Aliased Robot Manipulation},
  year                 = {2026},
  note                 = {{arXiv:}2605.14712},
  eprint               = {2605.14712},
  archiveprefix        = {arXiv},
  sourceentrytype      = {inproceedings},
  sourcebooktitle      = {Proceedings of the 2026 Conference on Empirical Methods in Natural Language Processing (EMNLP)},
  sourcenote           = {Accepted to EMNLP 2026; arXiv:2605.14712},
  sourceurl            = {https://arxiv.org/abs/2605.14712},
}

@misc{qwen2026qwen35,
  author               = {{Qwen Team}},
  title                = {{Qwen3.5}: Towards Native Multimodal Agents},
  howpublished         = {Qwen Blog},
  year                 = {2026},
  month                = {Feb.},
  url                  = {https://qwen.ai/blog?id=qwen3.5},
}

@inproceedings{hu2022lora,
  author               = {Hu, Edward J. and Shen, Yelong and Wallis, Phillip and Allen-Zhu, Zeyuan and Li, Yuanzhi and Wang, Shean and Wang, Lu and Chen, Weizhu},
  title                = {{LoRA}: Low-Rank Adaptation of Large Language Models},
  booktitle            = {Proc. Int. Conf. Learn. Represent. (ICLR)},
  year                 = {2022},
  url                  = {https://openreview.net/forum?id=nZeVKeeFYf9},
  sourcebooktitle      = {International Conference on Learning Representations},
}

@electronic{physicalintelligence2026openpi,
  author               = {{Physical Intelligence}},
  title                = {{``OpenPI: Open-source models and packages for robotics.''}\viaendwebtitle},
  howpublished         = {GitHub},
  note                 = {{Accessed:} Sep. 12, 2026},
  url                  = {https://github.com/Physical-Intelligence/openpi},
  urldate              = {2026-09-12},
  sourcetitle          = {{OpenPI}: Open-source models and packages for robotics},
  sourceyear           = {2026},
  sourcehowpublished   = {GitHub repository},
  sourcenote           = {Source code accessed September 12, 2026},
}

@misc{zhu2023fanuc,
  author               = {Zhu, Xinghao and Tian, Ran and Xu, Chenfeng and Huo, Mingxiao and Zhan, Wei and Tomizuka, Masayoshi and Ding, Mingyu},
  title                = {{Fanuc} Manipulation: A Dataset for Learning-based Manipulation with {FANUC Mate 200iD} Robot},
  howpublished         = {[Dataset]},
  year                 = {2023},
  url                  = {https://sites.google.com/berkeley.edu/fanuc-manipulation},
}

@inproceedings{chen2023playfusion,
  author               = {Chen, Lili and Bahl, Shikhar and Pathak, Deepak},
  title                = {{PlayFusion}: Skill Acquisition via Diffusion from Language-Annotated Play},
  booktitle            = {Proc. 7th Conf. Robot Learn. (CoRL)},
  volume               = {229},
  pages                = {2012--2029},
  year                 = {2023},
  sourcebooktitle      = {Proceedings of The 7th Conference on Robot Learning},
  sourcepublisher      = {PMLR},
  sourceseries         = {Proceedings of Machine Learning Research},
  sourceurl            = {https://proceedings.mlr.press/v229/chen23c.html},
}

@misc{liu2025fastumi100k,
  author               = {Liu, Kehui and Jia, Zhongjie and Li, Yang and {Zhaxizhuoma} and Chen, Pengan and Liu, Song and Liu, Xin and Zhang, Pingrui and Song, Haoming and Ye, Xinyi and Cao, Nieqing and Wang, Zhigang and Zeng, Jia and Wang, Dong and Ding, Yan and Zhao, Bin and Li, Xuelong},
  title                = {{FastUMI-100K}: Advancing Data-driven Robotic Manipulation with a Large-scale {UMI}-style Dataset},
  year                 = {2025},
  note                 = {{arXiv:}2510.08022},
  doi                  = {10.48550/arXiv.2510.08022},
  eprint               = {2510.08022},
  archiveprefix        = {arXiv},
  primaryclass         = {cs.RO},
  sourcehowpublished   = {arXiv preprint arXiv:2510.08022},
  sourceurl            = {https://arxiv.org/abs/2510.08022},
}

@misc{dass2023jacoplay,
  author               = {Dass, Shivin and Yapeter, Jullian and Zhang, Jesse and Zhang, Jiahui and Pertsch, Karl and Nikolaidis, Stefanos and Lim, Joseph J.},
  title                = {{CLVR Jaco Play} Dataset},
  howpublished         = {[Dataset]},
  year                 = {2023},
  url                  = {https://github.com/clvrai/clvr_jaco_play_dataset},
}

@inproceedings{fang2024rh20t,
  author               = {Fang, Hao-Shu and Fang, Hongjie and Tang, Zhenyu and Liu, Jirong and Wang, Chenxi and Wang, Junbo and Zhu, Haoyi and Lu, Cewu},
  title                = {{RH20T}: A Comprehensive Robotic Dataset for Learning Diverse Skills in One-Shot},
  booktitle            = {Proc. IEEE Int. Conf. Robot. Automat. (ICRA)},
  pages                = {653--660},
  year                 = {2024},
  sourcebooktitle      = {2024 IEEE International Conference on Robotics and Automation (ICRA)},
  sourcepublisher      = {IEEE},
  sourceurl            = {https://rh20t.github.io/},
}

@inproceedings{belkhale2023hydra,
  author               = {Belkhale, Suneel and Cui, Yuchen and Sadigh, Dorsa},
  title                = {{HYDRA}: Hybrid Robot Actions for Imitation Learning},
  booktitle            = {Proc. 7th Conf. Robot Learn. (CoRL)},
  volume               = {229},
  pages                = {2113--2133},
  year                 = {2023},
  sourcebooktitle      = {Proceedings of The 7th Conference on Robot Learning},
  sourcepublisher      = {PMLR},
  sourceseries         = {Proceedings of Machine Learning Research},
  sourceurl            = {https://proceedings.mlr.press/v229/belkhale23a.html},
}

@inproceedings{rosete2023tacorl,
  author               = {Rosete-Beas, Erick and Mees, Oier and Kalweit, Gabriel and Boedecker, Joschka and Burgard, Wolfram},
  title                = {Latent Plans for Task-Agnostic Offline Reinforcement Learning},
  booktitle            = {Proc. 6th Conf. Robot Learn. (CoRL)},
  volume               = {205},
  pages                = {1838--1849},
  year                 = {2023},
  sourcebooktitle      = {Proceedings of The 6th Conference on Robot Learning},
  sourcepublisher      = {PMLR},
  sourceseries         = {Proceedings of Machine Learning Research},
  sourceurl            = {https://proceedings.mlr.press/v205/rosete-beas23a.html},
}

@inproceedings{shah2023mutex,
  author               = {Shah, Rutav and Mart{\'i}n-Mart{\'i}n, Roberto and Zhu, Yuke},
  title                = {{MUTEX}: Learning Unified Policies from Multimodal Task Specifications},
  booktitle            = {Proc. 7th Conf. Robot Learn. (CoRL)},
  volume               = {229},
  pages                = {2663--2682},
  year                 = {2023},
  sourcebooktitle      = {Proceedings of The 7th Conference on Robot Learning},
  sourcepublisher      = {PMLR},
  sourceseries         = {Proceedings of Machine Learning Research},
  sourceurl            = {https://proceedings.mlr.press/v229/shah23b.html},
}

@misc{li2026rynnbrain11,
  author               = {Li, Kehan and Hou, Bohan and Zhu, Minghao and Zhang, Tianyi and Cheng, Zesen and Wang, Zhikai and Leng, Sicong and Li, Xin and Lin, Xiao and Yao, Biying and Zeng, Minghua and Liu, Jiangpin and Dang, Ronghao and Guo, Jiayan and Huang, Siteng and Zhao, Haoyu and Ping, Heng and Zhao, Yaxi and Zhao, Tong and Wang, Kexiang and Lu, Tong and Xue, Shengke and Tang, Jiahao and Wang, Yulei and Wang, Zejing and Gao, Jianwei and Lu, Shijian and Liu, Chengju and Yang, Jianfei and Chen, Mingxiu and Zhao, Deli},
  title                = {{RynnBrain 1.1}: Towards More Capable and Generalizable Embodied Foundation Model},
  year                 = {2026},
  note                 = {{arXiv:}2607.17977},
  doi                  = {10.48550/arXiv.2607.17977},
  eprint               = {2607.17977},
  archiveprefix        = {arXiv},
  primaryclass         = {cs.RO},
  sourcehowpublished   = {arXiv preprint arXiv:2607.17977},
  sourceurl            = {https://arxiv.org/abs/2607.17977},
}

@misc{tan2026robobrain25,
  author               = {Tan, Huajie and Zhou, Enshen and Li, Zhiyu and Xu, Yijie and Ji, Yuheng and Chen, Xiansheng and Chi, Cheng and Wang, Pengwei and Jia, Huizhu and Ao, Yulong and Cao, Mingyu and Chen, Sixiang and Li, Zhe and Liu, Mengzhen and Wang, Zixiao and Rong, Shanyu and Lyu, Yaoxu and Zhao, Zhongxia and Co, Peterson and Li, Yibo and Han, Yi and Xie, Shaoxuan and Yao, Guocai and Wang, Songjing and Zhang, Leiduo and Yang, Xi and Jiao, Yance and Shi, Donghai and Xie, Kunchang and Nie, Shaokai and Men, Chunlei and Lin, Yonghua and Wang, Zhongyuan and Huang, Tiejun and Zhang, Shanghang},
  title                = {{RoboBrain 2.5}: Depth in Sight, Time in Mind},
  year                 = {2026},
  note                 = {{arXiv:}2601.14352},
  doi                  = {10.48550/arXiv.2601.14352},
  eprint               = {2601.14352},
  archiveprefix        = {arXiv},
  primaryclass         = {cs.RO},
  sourcehowpublished   = {arXiv preprint arXiv:2601.14352},
  sourceurl            = {https://arxiv.org/abs/2601.14352},
}

@inproceedings{reimers2019sbert,
  author               = {Reimers, Nils and Gurevych, Iryna},
  title                = {{Sentence-BERT}: Sentence Embeddings using {Siamese BERT}-Networks},
  booktitle            = {Proc. Conf. Empirical Methods Natural Lang. Process. 9th Int. Joint Conf. Natural Lang. Process. (EMNLP-IJCNLP)},
  pages                = {3982--3992},
  year                 = {2019},
  note                 = {doi: {\urlstyle{same}\url{10.18653/v1/D19-1410}}},
  doi                  = {10.18653/v1/D19-1410},
  sourcebooktitle      = {Proceedings of the 2019 Conference on Empirical Methods in Natural Language Processing and the 9th International Joint Conference on Natural Language Processing},
  sourcepublisher      = {Association for Computational Linguistics},
  sourceurl            = {https://aclanthology.org/D19-1410/},
}

@inproceedings{zhai2023siglip,
  author               = {Zhai, Xiaohua and Mustafa, Basil and Kolesnikov, Alexander and Beyer, Lucas},
  title                = {Sigmoid Loss for Language Image Pre-Training},
  booktitle            = {Proc. IEEE/CVF Int. Conf. Comput. Vis. (ICCV)},
  pages                = {11975--11986},
  year                 = {2023},
  sourcebooktitle      = {Proceedings of the IEEE/CVF International Conference on Computer Vision},
  sourceurl            = {https://openaccess.thecvf.com/content/ICCV2023/html/Zhai_Sigmoid_Loss_for_Language_Image_Pre-Training_ICCV_2023_paper.html},
}

@electronic{rokokoStudio243,
  author               = {{Rokoko}},
  title                = {{``Rokoko Studio 2.4.3.''}\viaendwebtitle},
  howpublished         = {Rokoko Support},
  note                 = {{Accessed:} Sep. 8, 2026},
  url                  = {https://support.rokoko.com/hc/en-us/articles/21445198239761-Rokoko-Studio-2-4-3},
  urldate              = {2026-09-08},
  sourcetitle          = {{Rokoko Studio 2.4.3}},
  sourceyear           = {2024},
  sourcehowpublished   = {Rokoko release notes},
  sourcenote           = {Accessed September 8, 2026},
}

@misc{xie2026turbovla,
  author               = {Xie, Hengyi and Yao, Chenfei and Wu, Xianjin and Zhu, Yingying and Liang, Dingkang and Bai, Xiang and Ding, Han},
  title                = {{TurboVLA}: Real-Time Vision-Language-Action Model at 32 {Hz} on an {RTX} 4090 with $<$1 {GB} {VRAM}},
  year                 = {2026},
  note                 = {{arXiv:}2607.27205},
  doi                  = {10.48550/arXiv.2607.27205},
  eprint               = {2607.27205},
  archiveprefix        = {arXiv},
  sourcehowpublished   = {arXiv preprint arXiv:2607.27205},
  sourcenote           = {arXiv v2, 16 August 2026},
  sourceurl            = {https://arxiv.org/abs/2607.27205},
}

\end{document}